\documentclass[a4paper,twoside]{article}

\usepackage{epsfig}
\usepackage{subcaption}
\usepackage{calc}
\usepackage{amssymb}
\usepackage{amstext}
\usepackage{amsmath}
\usepackage{amsthm}
\usepackage{multicol}
\usepackage{pslatex}
\usepackage{apalike}

\usepackage[ruled,vlined, linesnumbered]{algorithm2e}
\usepackage{placeins}
\usepackage{multirow}

\usepackage{SCITEPRESS}     % Please add other packages that you may need BEFORE the SCITEPRESS.sty package.

\begin{document}
% My changes
\renewcommand{\abstractname}{Description}
\newtheorem{problem}{\textbf{Problem Definition}}
\SetKwInOut{Parameters}{Parameters}

\title{Acoustic Leak Detection in Water Networks}

\author{\authorname{Robert Müller\sup{1}\orcidAuthor{0000-0003-3108-713X}, 
Steffen Illium\sup{1}\orcidAuthor{0000-0003-0021-436X} Fabian Ritz\sup{1}\orcidAuthor{0000-0001-7707-1358}, Tobias Schröder\sup{2}, Christian Platschek\sup{2},\\ Jörg Ochs\sup{2} and
Claudia Linnhoff-Popien\sup{1}\orcidAuthor{0000-0001-6284-9286}}
\affiliation{\sup{1}Mobile and Distributed Systems Group, LMU Munich, Germany}
\email{\{robert.mueller, steffen.illium, fabian.ritz, linnhoff\}@ifi.lmu.de}
\affiliation{\sup{2}Stadtwerke M\"unchen GmbH, Germany}
\email{\{schroeder.tobias, platschek.christian, ochs.joerg\}@swm-infrastruktur.de}
}

\keywords{Leak Detection, Water Networks, Acoustic Anomaly Detection, Applied Machine Learning}

%\author{\IEEEauthorblockN{\IEEEauthorrefmark{1}Robert M\"uller, \IEEEauthorrefmark{1}Steffen %Illium, \IEEEauthorrefmark{1}Fabian Ritz, \\\IEEEauthorrefmark{2}Tobias Schr\"oder, %\IEEEauthorrefmark{2}Christian Platschek, \IEEEauthorrefmark{2}J\"org Ochs, %\IEEEauthorrefmark{1}Claudia Linnhoff-Popien}
%\IEEEauthorblockA{\IEEEauthorrefmark{1}\textit{Mobile and Distributed Systems Group, LMU Munich, %Germany} \\
%\IEEEauthorrefmark{2}\textit{Stadtwerke M\"unchen GmbH, Germany}\\
%\IEEEauthorrefmark{1}\texttt{\{robert.mueller,steffen.illium,fabian.ritz,linnhoff\}@ifi.lmu.de}\\
%\IEEEauthorrefmark{2}\texttt{\{schroeder.tobias,platschek.christian,ochs.joerg\}@swm-infrastruktur%.de}}
%}

%\maketitle

\abstract{
In this work, we present a general procedure for acoustic leak detection in water networks that satisfies multiple real-world constraints such as energy efficiency and ease of deployment. 
Based on recordings from seven contact microphones attached to the water supply network of a municipal suburb, we trained several shallow and deep anomaly detection models.
Inspired by how human experts detect leaks using electronic sounding-sticks, we use these models to repeatedly listen for leaks over a predefined decision horizon. This way we avoid constant monitoring of the system. While we found the detection of leaks in close proximity to be a trivial task for almost all models, neural network based approaches achieve better results at the detection of distant leaks. 
}
\onecolumn \maketitle \normalsize \setcounter{footnote}{0} \vfill

%\begin{IEEEkeywords}
%Leak Detection, Water Networks, Acoustic Anomaly Detection, Applied Machine Learning
%\end{IEEEkeywords}

\section{Introduction}
\label{sec:intro}

Leakage is one of the main causes for water loss in water supply and distribution networks.
Undetected leaks, which usually occur due to corrosion and soil movement, may have extensive negative effects on the surrounding infrastructure, customer convenience and financial profit. Contributing to this problem is the fact that it can take a considerable amount of time until a leak is detected, localized and countermeasures are put into place.\\
In the UK, approximately 3200 Million liters of water are wasted due to leakages in water networks every day~\cite{wateruk}, a disproportionately high value with regard to climate change and a shortage of drinking water in many countries. Greater efforts are needed to minimize water loss through leakages.\\
\begin{figure}
  \centering
  \includegraphics[width=1.\linewidth]{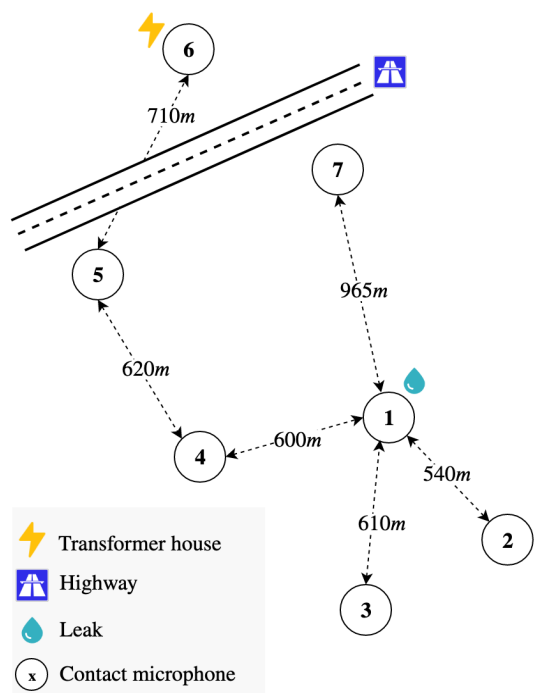}
  \caption{Simplified map of the location of each contact microphone and significant objects in their neighborhood.
  For each contact microphone, we depict the distance of the shortest pipeline link to another contact microphone. Note that distance is give n in meters of water pipe.}
  \label{fig:map}
\end{figure}
A reliable indicator for substantial water loss is the deviation of the zero-consumption status
in a balance area at night. But if a leakage is more subtle, it is usually not detected until water emerges from the surfaces and residents report the issue.\\ 
Locating the source by means of the acoustic emission from exiting water is one the predominantly used approaches~\cite{el2019leak}. The process is as follows:\\ 
\textit{1) Leak noise correlation:}
Liquid escaping a pipe creates shock waves that cause the pipe to vibrate, resulting in a characteristic leak sound. At least two sensors are attached to the pipe around the presumed leak location. By analyzing the variation in propagation times of the leak's acoustic emission between the sensors, the location is further narrowed down.
This approach requires infrastructural knowledge, such as pipe material, diameter, length and corresponding sound velocities.\\
\textit{2) Electro-acoustic method:} To further pinpoint the location, human leak detection experts search for leak noises using electronic sounding sticks. A leak can be localized by exploiting the fact that leak sounds become more dominant as the expert approaches the leak.
The process relies on trained human experts and requires a solid estimate of the leakage location as finding the first evidence might otherwise become very time consuming.\\
In this work, we aim to automatize the electro-acoustic method utilizing machine learning. We record normal operation data using seven contact microphones attached on various parts of the water supply network in a suburban area of Munich.
Subsequently, we train several unsupervised anomaly detection models and embed these into a 
broader procedure that satisfies several real world constraints, such as energy consumption and limited bandwidth. Our approach can be understood as a first step towards a fully automatized leak detection system that does not need to wait for leak aftereffects to appear. Moreover, our approach is data-driven and does not require a detailed mathematical model of the water network's inner workings.\\
The rest of the paper is structured as follows: In Section~\ref{sec:related_work} we briefly review related work. Then we motivate and propose a general procedure for acoustic leak detection in Section~\ref{sec:method}. In Section~\ref{sec:data_acquisition} we describe the data acquisition process followed by the description of the experimental setup and its evaluation. We close by summarizing our findings in Section~\ref{sec:conclusion} and laying out future work in Section~\ref{sec:future_work}
\section{Related Work}
\label{sec:related_work}
Related approaches of acoustic leak detection are mostly based on cross-correlation~\cite{muggleton2004,gao2017improving}, wavelet transforms~\cite{ni2002wavelet,ting2019improvement}, Support-Vector-Machines~\cite{kang2017novel,cody2017one} or neural networks~\cite{kang2017novel,chuang2019,cody2020detecting}. Other methods combine acoustic data with additional sensory measurements~\cite{Stoianov2007}. Their performance depends on the pipe network material (e.g. polyethylene or metal). However, none of these approaches directly matches the setup and constraints of this work as experiments are mostly carried out under laboratory conditions and only study a single algorithmic approach. Other physical phenomena that occur in presence of a leak can also be taken advantage of, e.g. using thermography, ground penetration radar or pressure-based methods.
An exhaustive summary can be found in~\cite{adedeji2017,chan2018}.\\
Apart from leak detection, the broader field of acoustic anomaly detection has recently gained traction.\\ Most of the recent work in the field is based upon Deep-Autoencoders (AEs). An AE learns to reconstruct its input from a compressed latent representation. Unseen, anomalous data is assumed to have a higher reconstruction error. The approaches mostly differ in the architecture used.\\
Duman et al.~\cite{duman2019acoustic} use a deep convolutional AE on the spectrograms of sounds from industrial processes. In~\cite{meire2019} the authors also conclude that convolutional AEs perform well on the task of acoustic anomaly detection while they have also found the One-Class Support Vector Machine to be a strong competitor. In the same vein, M\"uller et al.~\cite{mller2020acoustic} use pretrained convolutional neural networks for feature extraction and train various traditional anomaly detection models.
Other work~\cite{marchi2015novel,li2018anomalous,Nguyen2019} explicitly takes the sequential nature of sound into account by training a recurrent AE based on Long-Short-Term Memory.
Koizumi et al~\cite{koizumi2017optimizing} use a more traditional Feed-Forward AE in conjunction with a novel loss function based on statistical hypothesis testing. This approach requires the simulation of anomalous sounds by using rejection sampling.
Recently, Suefusa et al.~\cite{suefusa202} proposed to alter the input an AE receives to improve the detection of non-stationary sounds. Instead of predicting all spectrogram frames, they remove the center frame and use it as prediction target thereby alleviating the  difficulty of predicting the edge frames.\\
Another line of work investigates upon methods that operate directly on the raw waveform~\cite{Hayashi2018,Rushe2019}. WaveNet-like~\cite{Oord2016} generative architectures that utilize causal-dialated convolutions are used to predict the next sample. The prediction error serves as the anomaly score. 

\section{Motivation and Problem Definition}
\label{sec:system_requirements}
An automated leak detection system should be energy efficient, easy to deploy and easy to update.
We derive the problem definition from the following deployment scenario: Small battery driven IoT devices record sounds from contact microphones upon request from a central control unit. 
During a predefined period, sounds are periodically recorded for a short time frame (2-5s).
Afterwards, the collection of short audio sequences is transmitted to the central control unit via a low energy, low bandwidth radio network \cite{lora2018}. This reduces the high energy and bandwidth consumption that constant monitoring would cause. The central control unit can then put all received measurements in context to decide whether a leak is present or not. Moreover, using a central control unit makes updating the system (e.g. changing the algorithm or retraining the model) more feasible. This approach is inspired by how human leak detection experts make decisions using electronic sounding sticks. After attaching the stick on to a hydrant connection, the expert listens carefully for up to ten seconds. When facing audible noise emitted by infrastructure, agriculture or industry, the process is repeated until an accurate decision can be made. 
For precise leak localization, various hydrant or valve connections in the area are checked accordingly. Increasing leak sounds indicate a decreased distance to the leak. In this work, we focus on leak detection and leave localization for future work. Our approach is formulated as follows:\\

\begin{problem}
    \label{prob:prob_form}
    Let $X$ be some representation of an $h$ minute recording from a contact microphone on a water pipe. Further, let $\mathcal{F}_\theta: Y \xrightarrow{} \mathbb{R}_{+}$ be some trainable function with parameters $\theta$ where $Y$ is a sample of $X$ with a length of $t$ seconds. 
    Apply $\mathcal{F}_{\phi}$ on $m$ different sections of $X$ to obtain a vector $\in \mathbb{R}_{+}^{m}$ of positive real valued anomaly scores.
    To compute a single anomaly score for $X$, combine the measurements using an aggregation function $\phi: \mathbb{R}_{+}^{m} \xrightarrow{}  \mathbb{R}_{+}$. Find $(\mathcal{F}_\theta, \phi, h, t, m)$ such that the anomaly scores for leak sections are higher than the anomaly-scores for no-leak sections.\\
\end{problem}
\begin{algorithm}
\DontPrintSemicolon
\SetAlgoLined
\KwIn{Audio recording $X$}
\Parameters{
Score func. $\mathcal{F}_\theta$,
Aggregation func. $\phi$,
Number of samples $m$,
Sample length $t$,
preprocessing func. $\rho$
}
\KwOut{Anomaly Score for $X$ }
 \Begin{
    $h \longleftarrow \text{length}(X)$\;
    $P \longleftarrow \text{linspace}(0, h-t, m)$\;
    $S \longleftarrow [\;]$\;
    \For{$p$ in $P$}{
        $x \longleftarrow X[p:p+t]$\;
        $\Tilde{x} \longleftarrow \rho(x)$\;
        $s \longleftarrow \mathcal{F}_{\theta}(\Tilde{x})$\;
        append $s$ to $S$\;
    }
     \Return{$\phi(S)$}
 }
 \caption{Acoustic Leak Detection}
 \label{algo:eval}
\end{algorithm}
\begin{figure}
  \centering
  \includegraphics[width=1.0\linewidth]{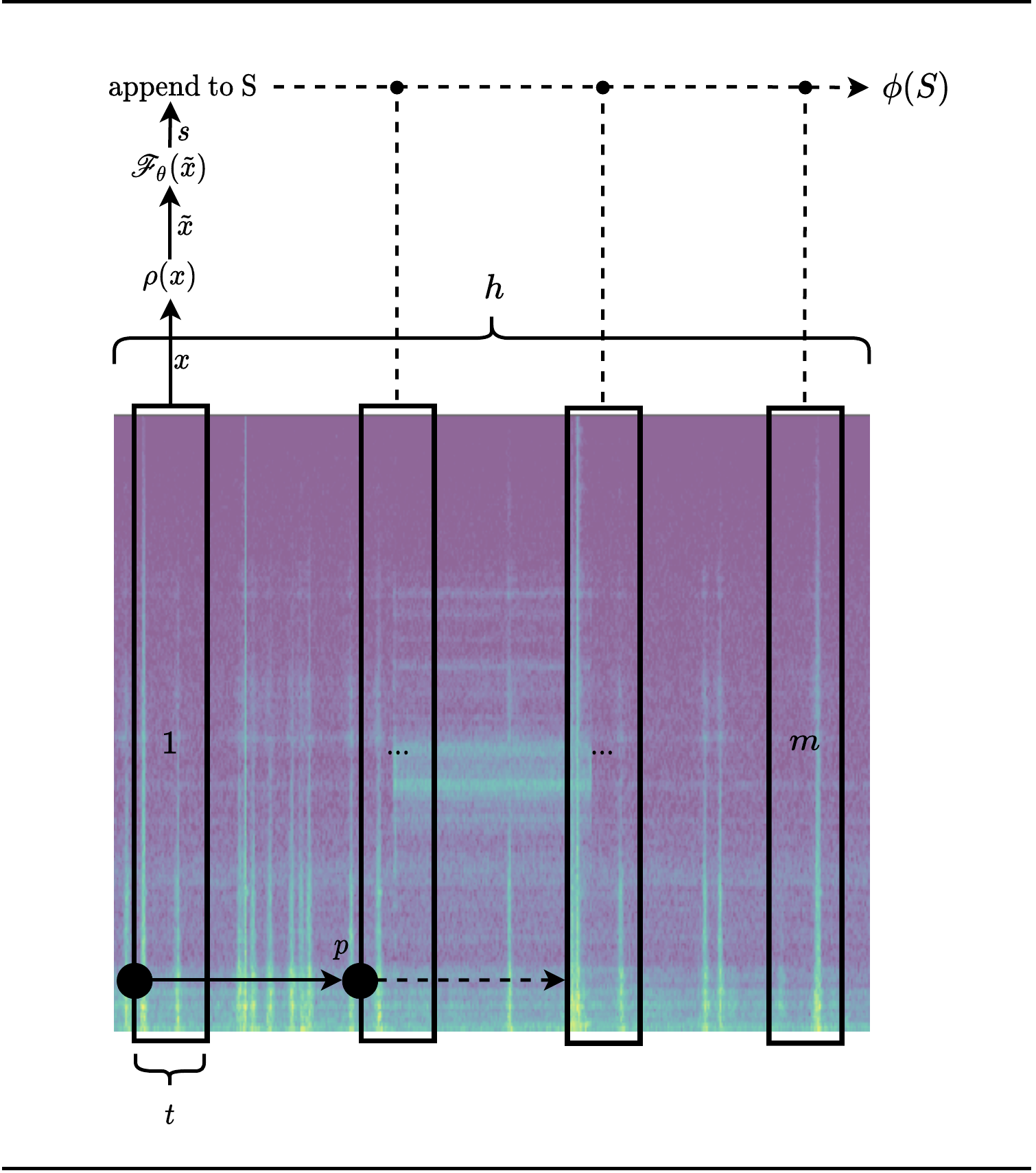}
  \caption{Visualization of the most important aspects of our acoustic leak detection approach.}
  \label{fig:method}
\end{figure}
The accompanying algorithm is depicted in Algorithm \ref{algo:eval}. Note that the sampling timepoints are equally distributed across the whole recording (Line 3), i.e. \texttt{linspace} returns the set $$\{\lfloor i*\frac{(h-t)}{m} \rfloor | i \in 1, \dots m\}$$ and each sample is preprocessed (Line 7) to fit the domain of the score function (e.g. spectrum, raw-audio or feature vector). We also depict the most important aspects of the method visually in Figure~\ref{fig:method} To obtain enough leak sound recordings for supervised learning, one would have to artificially create a vast amount of leaks. Furthermore, for these recordings to contain enough diversity, this would have to be done on various sections of the water network. Due to the financial and environmental issues that would arise, we assume that $\mathcal{F}_{\phi}$ is trained on normal operation data only.

%An fast and easy way to determine the leakage state of an pipe system is by its acoustic emissions.
%Traditionally, this is done by flow control sensors that monitor large network parts.
%If a leakage is assumed from those continuous measurements, experts in leak-detection have to "hear the pipe noise" by visiting the infrastructure at the spot.
%Leaking water and tailing water will then emit some kind of noise on movement and exit.
%Depending on the used material, these acoustic emissions propagate through the pipes physical structure.

% Einfach ganz viele Daten des normalfalls zu sammeln
% Lecks sind selten
% wie werden lecks aktuell entdeckt? -> Durchfluss an wenigen großen stellen, dann mensch losschicken der weiter eingrenzt

%From working in with the local service provider, the availability of energy in a modern city with a well maintained %infrastructure not as a critical resource rather then a given requirement. 
%On the other hand, information or data transfer in terms of available service quality and costs are of greater %interest. 

% Hence, we build a system that's not in need of power preservation, while trying to reduce the communication load and interval.

\section{Data Acquisition}
\label{sec:data_acquisition}
For data acquisition, a suburban part of Munich with little traffic was chosen as a testing ground. Seven contact microphones were directly attached onto hydrants of an iron pipe network ($\varnothing$ 100mm). % across the area.
Figure \ref{fig:map} shows a simplified map marking the locations of the microphones as well as significant objects in the surrounding  area that might influence the quality of the recordings. 
Sounds were recorded for three months. To avoid excessively high signals (overdrive), we kept the pre-amplifier, equalizer and post-amplifier neutral, effectively recording sounds "as-is" and disabling the dynamic sensitivity adjustment. This leads to the recordings having a headroom of 24 dB on average. All contact microphones are designed to record sounds reliably between 300Hz and 3000Hz. During the time of recording, a medium sized water leakage occurred in close proximity ($\approx$ 20m) to contact microphone one (see Figure \ref{fig:map}). The leak was present for 28 days and was fixed thereafter. Furthermore, we conducted various field tests. We simulated a leakage by opening a hydrant near contact microphone five and attached a contact microphone to hydrant connections between contact microphone four and five every 100 meters. Overall, we were neither able to hear the leak nor able to observe typical leak patterns in the frequency spectrograms beyond a distance of 600 meters.

\begin{figure*}
    \centering
  % trim=left bottom right top, clip
  % trim=110 35 180 55, clip 
  \includegraphics[width=\textwidth]{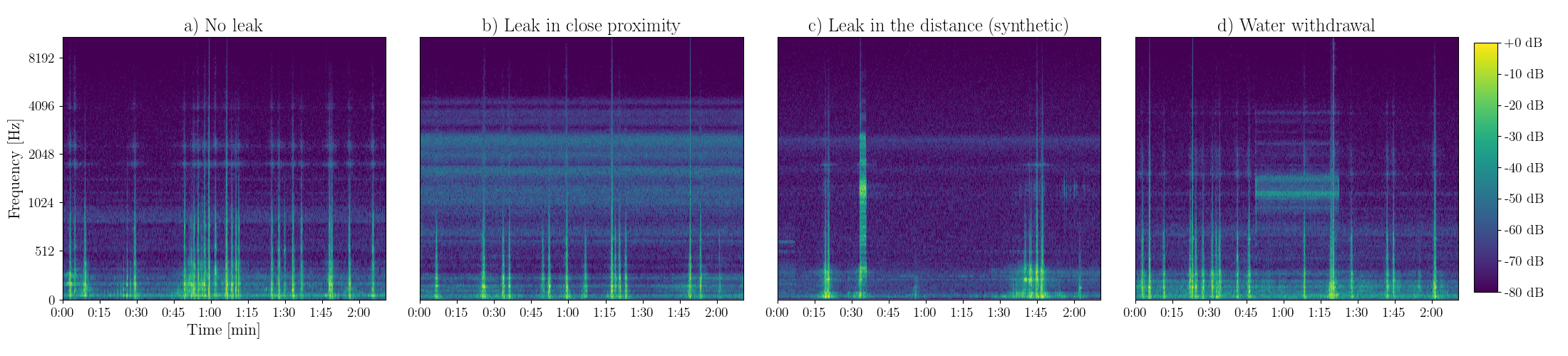}
  \caption{Four Mel-spectrograms depicting \textit{a)} normal operation, \textit{b)} a leak in close proximity, \textit{c)} a synthetically generated distant leak and \textit{d)} water withdrawal. 
  Spikes are caused by cars driving over the hydrant cover above the contact microphones.
  However, other recordings may also contain footsteps, animal noises or interfering noises due to a nearby transformer house. A leak is characterized by high energy in the upper frequencies. This pattern becomes less dominant as the distance to the leak increases and vanishes after $\approx 600m$. The leak shown in \textit{b)} has a throughput of $\approx 300\frac{\text{ml}}{\text{s}}$.
  }
  \label{fig:mel_spec_examples}
\end{figure*}

\section{Evaluation}
\label{sec:method}
In this section, we conduct various experiments to find an appropriate setting of $(\mathcal{F}_\theta, \phi, h, t, m)$. First we introduce the dataset followed by a brief discussion of the anomaly detection models that were used. Finally, we present the experiments and results.  
\subsection{Methodology and Dataset}
The evaluation set up is designed to line up with Problem Definition \ref{prob:prob_form}. We use 600 hours of normal operation recordings equally sampled from all contact microphones. 
Sounds were recorded in mono with a sampling rate of 16 kHz and a bit depth of 16 bit. A band-pass filter is applied to remove information above and below the contact microphone's supported frequencies.
For training, we split the data into $t=\{2s,\ 5s\}$ long audio samples with no overlap. These values are at the lower end of how long a human expert listens using a sounding stick. Samples are then preprocessed (Algorithm \ref{algo:eval}, Line  7) by either computing its mel-spectrogram or extracting eight spectral features (chromagram, spectral centroid, spectral bandwidth, spectral contrast, spectral roll off, spectral flatness, zero-crossing rate and the root-mean-square value) which we have identified to be good descriptors of a leak during early stages of research. Each feature-vector is standardized by subtracting the mean and dividing by the standard-deviation of the individual features. Figure \ref{fig:mel_spec_examples} provides more insights into the characteristics of the dataset. 
To evaluate the ability of our approach to differentiate between leak and no-leak, we use the following two datasets:\\

\textbf{Leak in close proximity:} To measure the ability to detect leaks in close proximity to a contact microphone, we select five consecutive days during which a leak was present in close proximity to contact microphone 1. For each day, we use the recordings from contact microphone 1 and one other contact microphone. This results in a balanced evaluation dataset with an equal number of leak and no-leak recordings.\\

\textbf{Leak in the distance (synthetic):} This setting measures how well a leak can be detected when it is farther away from a contact microphone. As we were only able to obtain recordings from a leakage near a single contact microphone, we mixed normal recordings with leak sounds having a high Signal-to-Noise Ratio of $+24\text{dB}$. Note that here noise stands for the leak sounds. Doing so yields synthetically generated recordings that resemble the characteristics of a leakage in the distance (Section \ref{sec:data_acquisition}). Synthetically creating anomalous recordings is a common approach when anomalous data is scarce~\cite{duman2019acoustic,purohit2019,koizumi2019,socoro2015,stowell2015,nakajima2016}.
We select $96$ hours of no-leak recordings from contact microphones two and three. $48$ hours of these recordings were mixed with randomly sampled leak sounds taken from a time-span between $0$a.m. - $4$a.m. This time-span was chosen as it yields pure leak sounds with very little noise.
\begingroup
\setlength{\tabcolsep}{6.5pt} % Default value: 6pt
\renewcommand{\arraystretch}{1.4} % Default value: 1
\begin{table*}[ht]
\small
\centering
\caption{Average AUCs with standard deviations (over 5 seeds) for different settings of $\mathcal{F}, h \text{ and } t$ where $\phi=\text{median}$, $m=20,h=30$min and $m=40,h=60$min. }
\begin{tabular}{lll|ll|llll}
\hline
                                          & \multicolumn{4}{c|}{Leak in close proximity}                                                                 & \multicolumn{4}{c}{Leak in the distance (synthetic)}                                                                               \\ \hline
\multicolumn{1}{l|}{Decision horizon $h$} & \multicolumn{2}{c|}{30min}                            & \multicolumn{2}{c|}{60min}                           & \multicolumn{2}{c|}{30min}                                                 & \multicolumn{2}{c}{60min}                             \\
\multicolumn{1}{l|}{Sample length $t$}    & 2s                        & 5s                        & 2s                        & 5s                       & 2s                        & \multicolumn{1}{l|}{5s}                        & 2s                        & 5s                        \\ \hline
\multicolumn{1}{l|}{GMM}                  & 74.8$\pm 2.5$             & 88.0$\pm 1.3$             & 74.6$\pm 2.1$             & 87.6$\pm 1.6$            & 64.0$\pm 5.5$             & \multicolumn{1}{l|}{68.7$\pm 0.0$}             & 68.4$\pm 1.0$             & 70.0$\pm 1.2$             \\
\multicolumn{1}{l|}{B-GMM}                & 78.7$\pm 3.7$             & 88.1$\pm 2.4$             & 78.8$\pm 4.0$             & 87.7$\pm 2.6$            & 64.0$\pm 5.1$             & \multicolumn{1}{l|}{70.5$\pm 1.8$}             & 69.3$\pm 1.2$             & 72.5$\pm 2.9$             \\
\multicolumn{1}{l|}{IF}                   & 98.6$\pm 0.0$             & 98.8$\pm 0.0$             & $\textbf{98.8}$$\pm 0.0$  & \textbf{100}$\pm 0.0$    & 41.4$\pm 0.0$             & \multicolumn{1}{l|}{42.5$\pm 1.8$}             & 41.0$\pm 1.5$             & 42.3$\pm 2.3$             \\
\multicolumn{1}{l|}{RealNVP}              & 98.6$\pm 2.0$             & 98.0$\pm 3.6$             & 99.0$\pm 1.3$             & 98.1$\pm 3.7$            & 75.2$\pm 2.7$             & \multicolumn{1}{l|}{\textbf{76.1}$\pm 2.5$}             & 77.1$\pm 2.8$             & \textbf{78.3}$\pm 3.1$    \\
\multicolumn{1}{l|}{DCAE}                 & 98.2$\pm 0.0$             & $98.9$$\pm 0.0$           & \textbf{99.8}$\pm 0.0$    & \textbf{100}$\pm 0.0$    & 75.1$\pm 1.5$             & \multicolumn{1}{l|}{73.4$\pm 5.0$}             & 76.1$\pm 2.1$             & 76.0$\pm 3.3$             \\
\multicolumn{1}{l|}{AAE}                  & \textbf{98.9}$\pm 0.0$    & 99.0$\pm 0.0$             & \textbf{99.8}$\pm 0.0$    & 99.8$\pm 0.0$            & 75.0$\pm 4.4$             & \multicolumn{1}{l|}{69.9$\pm 3.2$}             & 77.0$\pm 5.2$             & 71.6$\pm 2.4$             \\
\multicolumn{1}{l|}{AVB}                  & 98.1$\pm 0.5$             & \textbf{99.5}$\pm 0.1$             & 99.6$\pm 0.5$             & \textbf{100}$\pm 0.0$    & \textbf{76.9}$\pm 3.4$    & \multicolumn{1}{l|}{75.5$\pm 2.7$}             & \textbf{78.9}$\pm 3.7$    & 77.5$\pm 2.8$             \\
\hline 
% \multicolumn{1}{l|}{U-NET}                & 98.1$\pm 1.0$             & \textbf{99.6}$\pm 0.3$    & 99.6$\pm 0.8$             & \textbf{100}$\pm 0.0$    & 76.4$\pm 4.5$             & \multicolumn{1}{l|}{\textbf{76.4}$\pm 3.6$}    & 77.5$\pm 5.7$             & 78.2$\pm 5.1$             \\ \hline 
\end{tabular}
\label{tab:results}
\end{table*}
\endgroup

\subsection{Anomaly Detection Models}
\label{sec:models}
To compute anomaly scores for each individual sample (Algorithm \ref{algo:eval}, Line 8) we use various density estimation, ensemble models and deep neural networks. Models can be further subdivided according to the input they receive.\\

\textbf{Mel-spectrogram Input} A mel-spectrogram is a logarithmically scaled spectrogram to better align with how humans perceive sound. We compute the mel-sepctograms with $64$ mels, fft window = $2048$ frames, hop length = $512$ frames and normalize them to lie in the range $[0,1]$. We also experimented with other settings but have found all methods to be robust against these parameters.
The following deep neural networks are trained: 
\textit{i) Deep Convolutional Auto Encoder} (DCAE) A neural network that compresses the input into a low dimensional representation and then reconstructs the input from this representation. We use $[4, 16, 32]$ convolution filters with $2 \times 2$ max-pooling in-between each layer with ReLU~\cite{agarap2018deep} as activation function. We train for $100$ epochs with a batch size of $128$, a L$2$ weight penalty of $10^{-6}$ and optimize the AE using ADAM~\cite{kingma2014adam} with a learning rate of $0.0001$. For reconstruction, the inverse operations (deconvolution and up sampling) are applied in reverse order.
\textit{ii) Adversarial Auto Encoder} (AAE)~\cite{makhzani2015} Adversarially trained DCAE such that the latent space spanned by the bottleneck features matches the prior distribution $\mathcal{N}(0, I)$.  
\textit{iii) Adversarial Variational Bayes}~\cite{mescheder2017adversarial} Adversarially trained variational DCAE.\\

\textbf{Feature-Vector Input} Here we extract eight spectral features (see Section \ref{sec:method}) for each sample. The resulting feature vectors are used to train \textit{i) Gaussian Mixture Model} (GMM)   A density estimation algorithm that models the underlying probability distribution as a mixture of Gaussians. Parameters are estimated using expectation-maximization. We use $16$ mixture components with diagonal covariance matrices.  \textit{ii) Bayesian  Gaussian  Mixture  Model}  (B-GMM) In contrast to a GMM, this model is trained using variational inference. We use the same parameters as for the GMM. 
\textit{iii) RealNVP}~\cite{dinh2017} A sequence of invertible transformations modeled by a neural network that can directly compute the probability density of the data. The approach is based on normalizing flows. We use $3$ coupling layers, a hidden dimension of $150$, a batch size of $768$ and assume a normal distribution with zero mean and unit variance as base distribution.
All of the methods above use the log-probability of a sample as normality score.
\textit{iv) Isolation Forest} (IF)~\cite{liu2008} Recursively partitions the feature space. The number of splits needed to isolate a data point is used as normality score. We use $120$ base estimators.\\

Model performance is measured with the \textit{Area Under the Receiver Operating Characteristics} (AUC) which quantifies how well a model can distinguish between leak and no-leak across all classification thresholds. The AUC is the standard metric to evaluate anomaly detection models across many domains~\cite{aggarwal2015outlier,chalapathy2019deep} as it yields a complete sensitivity/specificity report. The evaluation setup follows~\cite{Ruff2018}.

\subsection{Choosing the Model}
The most important aspect of our approach is to choose a suitable score function $\mathcal{F}_{\phi}$. The datasets presented in Section \ref{sec:method} are split into consecutive $h=30$min and $h=60$min long recordings. On each of these recordings, we run Algorithm \ref{algo:eval} independently to obtain a single anomaly score. Here we chose the median as aggregation function $\phi$ due to its robustness against outliers\footnote{The median showed the best results compared to other aggregation functions like the mean. Using only the most normal sample (min-pooling) leads to a comparable, but worse performance.} and evaluate across all models from Section \ref{sec:models} with $t=2$ and $t=5$.
We set $m=20$ and $m=40$ for $h=30$ min and $h=60$ min, respectively. 
Model parameters were determined using a small development set.\\
Results are depicted in Table \ref{tab:results}.
In the case of \textit{Leak in close proximity} the performance of GMM and B-GMM is significantly worse compared to all other models. Interestingly, AUC-scores for GMM and B-GMM increase by approximately $12\%$ when $t=5$. This finding carries over the second setting as well. 
When $h=60$ and $t=5$  IF, DCAE and AVB reach perfect scores. Generally, the setting can be considered trivial for IF, RealNVP, DCAE, AVB and AAE. Note that in case of the AAE and AVB we also tried using the log-probability of the latent representation but the reconstruction error turned out to yield better results.\\
Results for \textit{Leak in the distance (synthetic)} paint a different picture. 
In this setting, the differences between inlier and outlier are more subtle and therefore considerably harder to detect. 
IF fails completely on this task due to the reduced distance between inlier and outlier in feature space. The number of splits is the same for almost all samples and mostly depends on the random splitting points. 
Moreover, we observe general superiority of the neural network (NN) based methods RealNVP, DCAE, AVB and AAE indicating that NNs better reveal the more subtle differences.
On $2$s, Mel-spectrogram based AVB performs best and on $5$s RealNVP outperforms all other methods. We observed that auto encoders work best with a small bottleneck as this limits their ability to generalize over leak patterns based on water withdrawal. Moreover, we can verify that the extracted feature set is indeed a good leak indicator as ist shows strong performance when used in conjunction with NN based RealNVP.

% However, results indicate a slight advantage of the compact feature-vector representation over mel-spectrogram based models when combined with density estimation using RealNVP. We argue that through higher expressiveness of neural networks, they are better able to capture even small deviations from the norm.

\subsection{Choosing the Number of Samples}
\begin{figure}
  \centering
  % trim=left bottom right top, clip
  % trim=110 35 180 55, clip 
  \includegraphics[width=1.\linewidth,trim=0 20 0 0, clip]{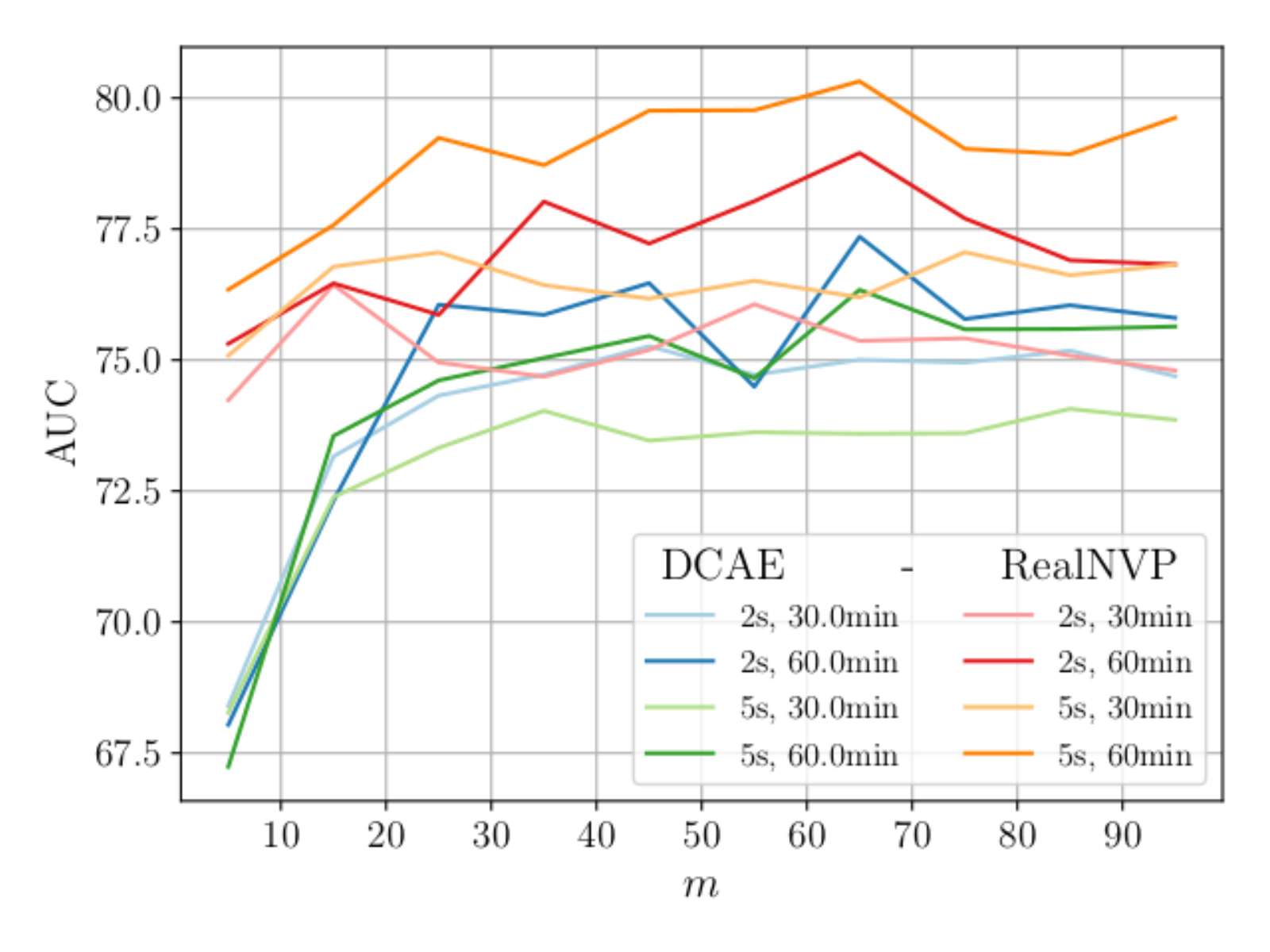}
    \caption{Performance comparison for $m=\{5, 15, 25, \dots 105\}$ on \textit{Leak in the distance (synthetic)}. Each value is the average AUC over $5$ seeds.}
  \label{fig:sampling_eval}
\end{figure}

In this experiment, we evaluate upon the influence of the number of samples on the model performance. In figure \ref{fig:sampling_eval}, we vary the number of samples $m$ from $5$ to $105$ in steps of $10$ samples. We evaluate across all combinations of decision horizon $h \in \{30\text{min},\ 60\text{min}\}$ and sample length $t \in \{2\text{s}, 5\text{s}\}$ on \textit{Leak in the distance (synthetic)} with $\phi$=median. For better clarity, we only show results for RealNVP and DCAE.\\
In Figure \ref{fig:sampling_eval}, we observe that using less than $15$ samples yields significantly worse performance. Results stabilize thereafter and peak performance is achieved at $m=65$ for most settings. We conclude that increasing the number of samples has a positive effect on the performance because it makes the decision less dependent on individual samples. More samples better represent the underlying distribution as a leak is characterized by a constantly present acoustic pattern.
In contrast to a streaming approach, only a fraction of all possible measurements has to be considered as we did not observe a substantial increase in performance by considering even more samples.

\section{Conclusion}
\label{sec:conclusion}
In this work, we presented a general procedure for leak detection in water networks that satisfies real-world constraints and provided a thorough evaluation of different parameter settings. While a leak in close proximity to a contact microphone is trivial for most scoring functions, neural network based approaches yielded superior results with respect to the detection of a (synthetic) leak in the distance. Additionally, we found that it is not necessary to constantly monitor the system. It suffices to consider a fraction of the recordings during a predefined decision horizon. 

\section{Future Work}
\label{sec:future_work}
Future work might investigate further upon different scoring functions, e.g. by taking the sequential nature of the recordings into account. 
Other avenues worth exploring are more sophisticated sampling and aggregation strategies. The extension of our approach to leak localization (e.g. via trilateration) represents the next logical step. Moreover, instead of training a single model on data from all contact-microphones, one might train separate models. Another possibility would be to use weight sharing and condition~\cite{Huang2017} the model on the contact-microphone ID or on features of their surrounding.
Another important aspect that still remains to be investigated, is how to update the model when new data (possibly collected in another area) arrives~\cite{Koizumi2020}.
\section*{Acknowledgments}
\addcontentsline{toc}{section}{Acknowledgments}
This work is part of the research project \textit{ErLoWa} which was carried out in cooperation with Stadtwerke M\"unchen GmbH.

%\bibliographystyle{IEEEtran}
%\bibliography{IEEEabrv,bibliography}
\bibliographystyle{apalike}
{\small
\bibliography{bibliography}}

\begin{thebibliography}{}

\bibitem[Adedeji et~al., 2017]{adedeji2017}
Adedeji, K.~B., Hamam, Y., Abe, B.~T., and Abu-Mahfouz, A.~M. (2017).
\newblock Towards achieving a reliable leakage detection and localization
  algorithm for application in water piping networks: An overview.
\newblock {\em IEEE Access}, 5:20272--20285.

\bibitem[Agarap, 2018]{agarap2018deep}
Agarap, A.~F. (2018).
\newblock Deep learning using rectified linear units (relu).
\newblock {\em arXiv preprint arXiv:1803.08375}.

\bibitem[Aggarwal, 2015]{aggarwal2015outlier}
Aggarwal, C.~C. (2015).
\newblock Outlier analysis.
\newblock In {\em Data mining}, pages 237--263. Springer.

\bibitem[Chalapathy and Chawla, 2019]{chalapathy2019deep}
Chalapathy, R. and Chawla, S. (2019).
\newblock Deep learning for anomaly detection: A survey.
\newblock {\em arXiv preprint arXiv:1901.03407}.

\bibitem[Chan et~al., 2018]{chan2018}
Chan, T., Chin, C.~S., and Zhong, X. (2018).
\newblock Review of current technologies and proposed intelligent methodologies
  for water distributed network leakage detection.
\newblock {\em IEEE Access}, 6:78846--78867.

\bibitem[Chuang et~al., 2019]{chuang2019}
Chuang, W.-Y., Tsai, Y.-L., and Wang, L.-H. (2019).
\newblock Leak detection in water distribution pipes based on cnn with mel
  frequency cepstral coefficients.
\newblock In {\em Proceedings of the 2019 3rd International Conference on
  Innovation in Artificial Intelligence}, pages 83--86.

\bibitem[Cody et~al., 2017]{cody2017one}
Cody, R., Narasimhan, S., and Tolson, B. (2017).
\newblock {One Class SVM – Leak Detection in Water Distribution Systems'}.

\bibitem[Cody et~al., 2020]{cody2020detecting}
Cody, R.~A., Tolson, B.~A., and Orchard, J. (2020).
\newblock Detecting leaks in water distribution pipes using a deep autoencoder
  and hydroacoustic spectrograms.
\newblock {\em Journal of Computing in Civil Engineering}, 34(2).

\bibitem[Dinh et~al., 2017]{dinh2017}
Dinh, L., Sohl{-}Dickstein, J., and Bengio, S. (2017).
\newblock Density estimation using real {NVP}.
\newblock In {\em 5th International Conference on Learning Representations,
  {ICLR} 2017}.

\bibitem[Duman et~al., 2019]{duman2019acoustic}
Duman, T.~B., Bayram, B., and {\.I}nce, G. (2019).
\newblock Acoustic anomaly detection using convolutional autoencoders in
  industrial processes.
\newblock In {\em International Workshop on Soft Computing Models in Industrial
  and Environmental Applications}, pages 432--442. Springer.

\bibitem[El-Zahab and Zayed, 2019]{el2019leak}
El-Zahab, S. and Zayed, T. (2019).
\newblock Leak detection in water distribution networks: an introductory
  overview.
\newblock {\em Smart Water}, 4(1):5.

\bibitem[Gao et~al., 2017]{gao2017improving}
Gao, Y., Brennan, M.~J., Liu, Y., Almeida, F.~C., and Joseph, P.~F. (2017).
\newblock Improving the shape of the cross-correlation function for leak
  detection in a plastic water distribution pipe using acoustic signals.
\newblock {\em Applied Acoustics}, 127:24--33.

\bibitem[Hayashi et~al., 2018]{Hayashi2018}
Hayashi, T., Komatsu, T., Kondo, R., Toda, T., and Takeda, K. (2018).
\newblock Anomalous sound event detection based on wavenet.
\newblock In {\em 2018 26th European Signal Processing Conference (EUSIPCO)},
  pages 2494--2498. IEEE.

\bibitem[Huang and Belongie, 2017]{Huang2017}
Huang, X. and Belongie, S. (2017).
\newblock Arbitrary style transfer in real-time with adaptive instance
  normalization.
\newblock In {\em Proceedings of the IEEE International Conference on Computer
  Vision}, pages 1501--1510.

\bibitem[Kang et~al., 2017]{kang2017novel}
Kang, J., Park, Y.-J., Lee, J., Wang, S.-H., and Eom, D.-S. (2017).
\newblock Novel leakage detection by ensemble cnn-svm and graph-based
  localization in water distribution systems.
\newblock {\em IEEE Transactions on Industrial Electronics}, 65(5):4279--4289.

\bibitem[Kingma and Ba, 2014]{kingma2014adam}
Kingma, D.~P. and Ba, J. (2014).
\newblock Adam: A method for stochastic optimization.
\newblock {\em arXiv preprint arXiv:1412.6980}.

\bibitem[Koizumi et~al., 2017]{koizumi2017optimizing}
Koizumi, Y., Saito, S., Uematsu, H., and Harada, N. (2017).
\newblock Optimizing acoustic feature extractor for anomalous sound detection
  based on neyman-pearson lemma.
\newblock In {\em 2017 25th European Signal Processing Conference (EUSIPCO)},
  pages 698--702. IEEE.

\bibitem[Koizumi et~al., 2019]{koizumi2019}
Koizumi, Y., Saito, S., Uematsu, H., Harada, N., and Imoto, K. (2019).
\newblock Toyadmos: A dataset of miniature-machine operating sounds for
  anomalous sound detection.
\newblock In {\em 2019 IEEE Workshop on Applications of Signal Processing to
  Audio and Acoustics (WASPAA)}, pages 313--317. IEEE.

\bibitem[Koizumi et~al., 2020]{Koizumi2020}
Koizumi, Y., Yasuda, M., Murata, S., Saito, S., Uematsu, H., and Harada, N.
  (2020).
\newblock Spidernet: Attention network for one-shot anomaly detection in
  sounds.
\newblock In {\em ICASSP 2020-2020 IEEE International Conference on Acoustics,
  Speech and Signal Processing (ICASSP)}, pages 281--285. IEEE.

\bibitem[Li et~al., 2018]{li2018anomalous}
Li, Y., Li, X., Zhang, Y., Liu, M., and Wang, W. (2018).
\newblock Anomalous sound detection using deep audio representation and a blstm
  network for audio surveillance of roads.
\newblock {\em IEEE Access}, 6:58043--58055.

\bibitem[Liu et~al., 2008]{liu2008}
Liu, F.~T., Ting, K.~M., and Zhou, Z.-H. (2008).
\newblock Isolation forest.
\newblock In {\em 2008 Eighth IEEE International Conference on Data Mining},
  pages 413--422. IEEE.

\bibitem[Makhzani et~al., 2015]{makhzani2015}
Makhzani, A., Shlens, J., Jaitly, N., Goodfellow, I., and Frey, B. (2015).
\newblock Adversarial autoencoders.

\bibitem[Marchi et~al., 2015]{marchi2015novel}
Marchi, E., Vesperini, F., Eyben, F., Squartini, S., and Schuller, B. (2015).
\newblock A novel approach for automatic acoustic novelty detection using a
  denoising autoencoder with bidirectional lstm neural networks.
\newblock In {\em 2015 IEEE International Conference on Acoustics, Speech and
  Signal Processing (ICASSP)}, pages 1996--2000. IEEE.

\bibitem[Meire and Karsmakers, 2019]{meire2019}
Meire, M. and Karsmakers, P. (2019).
\newblock Comparison of deep autoencoder architectures for real-time acoustic
  based anomaly detection in assets.
\newblock In {\em 2019 10th IEEE International Conference on Intelligent Data
  Acquisition and Advanced Computing Systems: Technology and Applications
  (IDAACS)}, volume~2, pages 786--790. IEEE.

\bibitem[Mescheder et~al., 2017]{mescheder2017adversarial}
Mescheder, L., Nowozin, S., and Geiger, A. (2017).
\newblock Adversarial variational bayes: Unifying variational autoencoders and
  generative adversarial networks.
\newblock In {\em Proceedings of the 34th International Conference on Machine
  Learning-Volume 70}, pages 2391--2400. JMLR. org.

\bibitem[Muggleton and Brennan, 2004]{muggleton2004}
Muggleton, J. and Brennan, M. (2004).
\newblock Leak noise propagation and attenuation in submerged plastic water
  pipes.
\newblock {\em Journal of Sound and Vibration}, 278(3):527--537.

\bibitem[M{\"u}ller et~al., 2020]{mller2020acoustic}
M{\"u}ller, R., Ritz, F., Illium, S., and Linnhoff-Popien, C. (2020).
\newblock Acoustic anomaly detection for machine sounds based on image transfer
  learning.
\newblock {\em arXiv preprint arXiv:2006.03429}.

\bibitem[Nakajima et~al., 2016]{nakajima2016}
Nakajima, Y., Naito, T., Sunago, N., Ohshima, T., and Ono, N. (2016).
\newblock Dnn-based environmental sound recognition with real-recorded and
  artificially-mixed training data.
\newblock In {\em INTER-NOISE and NOISE-CON Congress and Conference
  Proceedings}, volume 253, pages 1832--1841. Institute of Noise Control
  Engineering.

\bibitem[Nguyen et~al., 2019]{Nguyen2019}
Nguyen, D., Kirsebom, O.~S., Fraz{\~a}o, F., Fablet, R., and Matwin, S. (2019).
\newblock Recurrent neural networks with stochastic layers for acoustic novelty
  detection.
\newblock In {\em ICASSP 2019-2019 IEEE International Conference on Acoustics,
  Speech and Signal Processing (ICASSP)}, pages 765--769. IEEE.

\bibitem[Ni and Iwamoto, 2002]{ni2002wavelet}
Ni, Q.-Q. and Iwamoto, M. (2002).
\newblock Wavelet transform of acoustic emission signals in failure of model
  composites.
\newblock {\em Engineering Fracture Mechanics}, 69(6):717--728.

\bibitem[Oord et~al., 2016]{Oord2016}
Oord, A. v.~d., Dieleman, S., Zen, H., Simonyan, K., Vinyals, O., Graves, A.,
  Kalchbrenner, N., Senior, A., and Kavukcuoglu, K. (2016).
\newblock Wavenet: A generative model for raw audio.
\newblock {\em arXiv preprint arXiv:1609.03499}.

\bibitem[Purohit et~al., 2019]{purohit2019}
Purohit, H., Tanabe, R., Ichige, K., Endo, T., Nikaido, Y., Suefusa, K., and
  Kawaguchi, Y. (2019).
\newblock Mimii dataset: Sound dataset for malfunctioning indsutrial machine
  investigation and inspection.
\newblock In {\em Acoustic Scenes and Events 2019 Workshop (DCASE2019)}, page
  209.

\bibitem[Ruff et~al., 2018]{Ruff2018}
Ruff, L., Vandermeulen, R., Goernitz, N., Deecke, L., Siddiqui, S.~A., Binder,
  A., M{\"u}ller, E., and Kloft, M. (2018).
\newblock Deep one-class classification.
\newblock In {\em International Conference on Machine Learning}, pages
  4393--4402.

\bibitem[{Rushe} and {Namee}, 2019]{Rushe2019}
{Rushe}, E. and {Namee}, B.~M. (2019).
\newblock Anomaly detection in raw audio using deep autoregressive networks.
\newblock In {\em ICASSP 2019 - 2019 IEEE International Conference on
  Acoustics, Speech and Signal Processing (ICASSP)}, pages 3597--3601.

\bibitem[Socor{\'o} et~al., 2015]{socoro2015}
Socor{\'o}, J.~C., Ribera, G., Sevillano, X., and Al{\'\i}as, F. (2015).
\newblock Development of an anomalous noise event detection algorithm for
  dynamic road traffic noise mapping.
\newblock In {\em Proceedings of the 22nd International Congress on Sound and
  Vibration (ICSV22), Florence, Italy}, pages 12--16.

\bibitem[{Stoianov} et~al., 2007]{Stoianov2007}
{Stoianov}, I., {Nachman}, L., {Madden}, S., {Tokmouline}, T., and {Csail}, M.
  (2007).
\newblock Pipenet: A wireless sensor network for pipeline monitoring.
\newblock In {\em 2007 6th International Symposium on Information Processing in
  Sensor Networks}, pages 264--273.

\bibitem[Stowell et~al., 2015]{stowell2015}
Stowell, D., Giannoulis, D., Benetos, E., Lagrange, M., and Plumbley, M.~D.
  (2015).
\newblock Detection and classification of acoustic scenes and events.
\newblock {\em IEEE Transactions on Multimedia}, 17(10):1733--1746.

\bibitem[Suefusa et~al., 2020]{suefusa202}
Suefusa, K., Nishida, T., Purohit, H., Tanabe, R., Endo, T., and Kawaguchi, Y.
  (2020).
\newblock Anomalous sound detection based on interpolation deep neural network.
\newblock In {\em ICASSP 2020-2020 IEEE International Conference on Acoustics,
  Speech and Signal Processing (ICASSP)}, pages 271--275. IEEE.

\bibitem[SWM, 2018]{lora2018}
SWM, P. (2018).
\newblock Die swm vernetzen m{\"u}nchen: Lora-netz am start f{\"u}r das
  internet der dinge.
\newblock {https://www.swm.de/dam/swm/pressemitteilungen/
  2018/06/20180604-swm-bauen-lora-netz-auf.pdf}.

\bibitem[Ting et~al., 2019]{ting2019improvement}
Ting, L., Tey, J., Tan, A., King, Y., and Faidz, A. (2019).
\newblock Improvement of acoustic water leak detection based on dual tree
  complex wavelet transform-correlation method.
\newblock In {\em IOP Conference Series: Earth and Environmental Science},
  volume 268. IOP Publishing.

\bibitem[WaterUK, 2020]{wateruk}
WaterUK (2020).
\newblock Leaking pipes.
\newblock https://discoverwater.co.uk/leaking-pipes.
\newblock Accessed: 2020-01-20.

\end{thebibliography}

\end{document}